\documentclass{article}
\usepackage{ijcai25}
\usepackage{adjustbox}
\usepackage{times}
\usepackage{soul}
\usepackage{url}
\usepackage[hidelinks]{hyperref}
\usepackage[utf8]{inputenc}
\usepackage[small]{caption}
\usepackage{graphicx}
\usepackage{amsmath}
\usepackage{amsthm}
\usepackage{booktabs}
\usepackage{algorithm}
\usepackage{algorithmic}
\usepackage[switch]{lineno}
\usepackage{placeins}
\usepackage{arydshln}
\usepackage{comment}
\usepackage{afterpage}

\title{Hamming Attention Distillation: Binarizing Keys and Queries for Efficient Long-Context Transformers}

\author{
Mark Horton$^1$
\and
Tergel Molom-Ochir$^1$\and
Peter Liu$^1$\and
Bhavna Gopal$^1$\and
Chiyue Wei$^1$\and
\\Cong Guo$^1$\and
Brady Taylor$^1$\and
Deliang Fan$^2$\and
Shan X. Wang$^3$\and
Hai Li$^1$\and
\And
Yiran Chen$^1$
\affiliations
$^1$Duke University\\
$^2$Arizona State University\\
$^3$Stanford University\\
\emails
\{mark.horton, tergel.molom-ochir, peter.liu, bhavna.gopal, chiyue.wei, cong.guo, brady.g.taylor, hai.li, yiran.chen\}@duke.edu,
dfan@asu.edu,
sxwang@stanford.edu
}

\begin{document}

\maketitle

\begin{abstract}
Pre-trained transformer models with extended context windows are notoriously expensive to run at scale, often limiting real-world deployment due to their high computational and memory requirements. In this paper, we introduce Hamming Attention Distillation (HAD), a novel framework that binarizes keys and queries in the attention mechanism to achieve significant efficiency gains. By converting keys and queries into {-1, +1} vectors and replacing dot-product operations with efficient Hamming distance computations, our method drastically reduces computational overhead. Additionally, we incorporate attention matrix sparsification to prune low-impact activations, which further reduces the cost of processing long-context sequences.
\par Despite these aggressive compression strategies, our distilled approach preserves a high degree of representational power, leading to substantially improved accuracy compared to prior transformer binarization methods. We evaluate HAD on a range of tasks and models, including the GLUE benchmark, ImageNet, and QuALITY, demonstrating state-of-the-art performance among binarized Transformers while drastically reducing the computational costs of long-context inference. \par We implement HAD in custom hardware simulations, demonstrating superior performance characteristics compared to a custom hardware implementation of standard attention.  HAD achieves just $\mathbf{1.78}\%$ performance losses on GLUE compared to $9.08\%$ in state-of-the-art binarization work, and $\mathbf{2.5}\%$ performance losses on ImageNet compared to $12.14\%$, all while targeting custom hardware with a $\mathbf{79}\%$ area reduction and $\mathbf{87}\%$ power reduction compared to its standard attention counterpart. 
\end{abstract}

\section{Introduction}

\begin{figure}[t]
    \centering
    \includegraphics[width=0.9\linewidth]{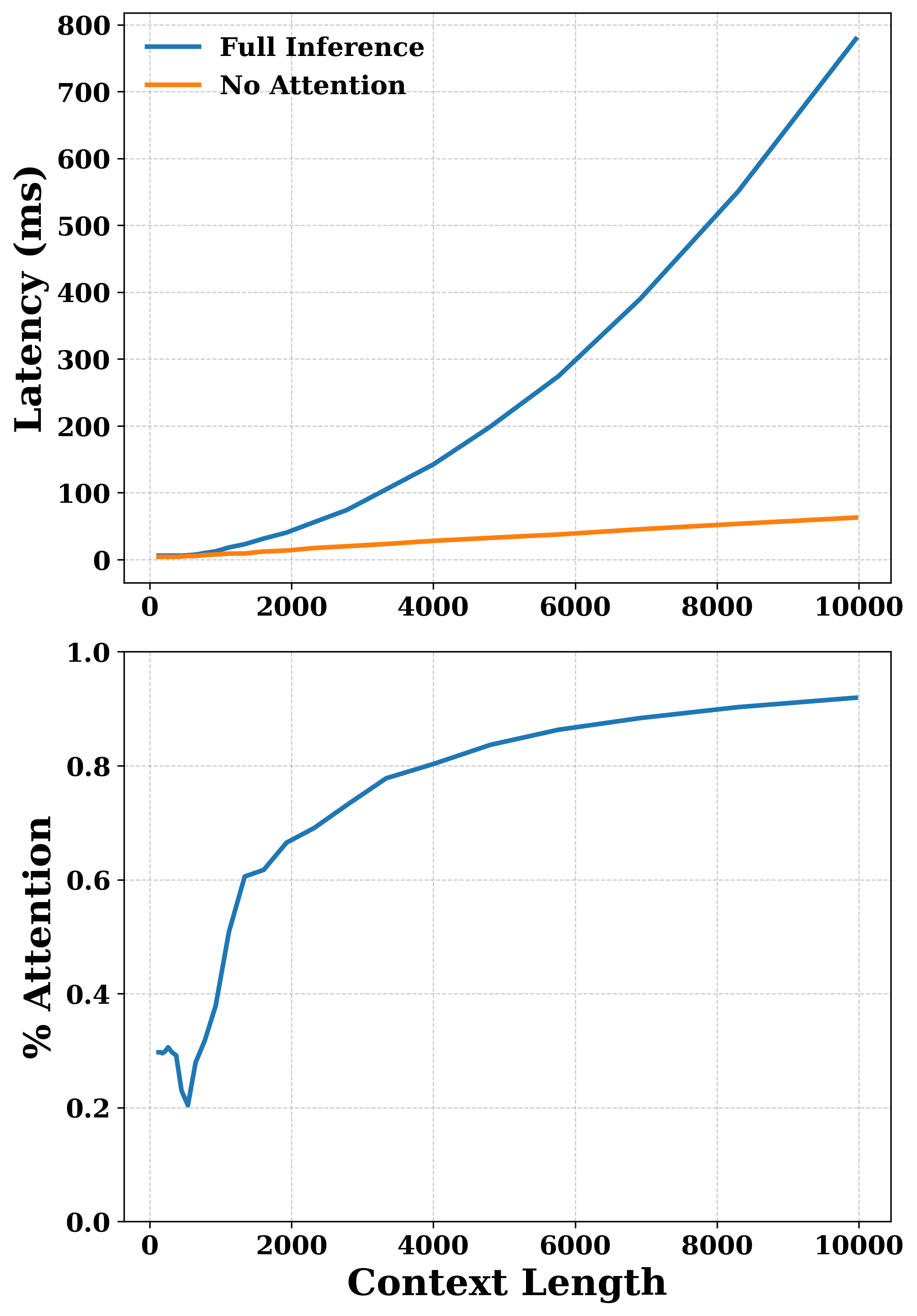}
    \caption{Runtime analysis of BERT Base over increasing context lengths on NVIDIA L40 GPU. As context length rises into the thousands, attention begins to dominate the runtime.  The top plot shows the latency of BERT inferenced with and without its attention, and the bottom shows the percentage of latency attributable to attention versus all other operations.}
    \label{fig:latency}
\end{figure}

Since the introduction of the transformer \cite{transformer}, various adaptations of the architecture have spread across nearly every domain of deep learning, achieving state-of-the-art performance across modalities such as language, vision, video, and audio \cite{bert,vit,video,audio}.  The versatility and success of the transformer is largely attributable to its self-attention module, which learns pairwise relationships and information sharing between vectors in its set of inputs, each of which can represent language tokens, image patches, etc.  While this provides a flexible approach to learn relationships in complex data, the pairwise operations also result in $O(n^2)$ runtime, where n is the number of inputs to the self-attention module.  This means that when processing a text, we would expect the runtime to scale with the square of the length of the text, while all other operations in the transformer scale linearly to the length of the text as visualized in Figure \ref{fig:latency}.  
\par This poor scaling is a critical bottleneck across many fields where we may want to apply transformers.  Chatbots processing very long texts for question answering, LLM-based search tools processing a large number of web pages, vision transformers applied to large and high-resolution images, and video models all demand context lengths past what standard transformers can efficiently accommodate.  This has led to a large research literature on alternatives for self-attention with better asymptotic scaling, targeting long-context settings \cite{mamba,bigbird,linformer}.  However, none of these approaches has overtaken standard self-attention in popularity for a number of reasons from accuracy/performance losses to poor hardware utilization and throughput.  Additionally, many of these approaches cannot be easily used to adapt an existing pre-trained transformer for long-context inference.
\par Binarization, a special case of quantization, is another research area which has sought to improve the runtime and performance characteristics of deep neural networks.  However, unlike the long context literature which focuses on asymptotic runtime advantages, the binarization literature focuses on efficiently leveraging hardware.  This is done by replacing floating point weights and activations with binary values, therefore replacing expensive floating point operations with far more memory and runtime efficient bitwise operations.  These approaches often target custom hardware \cite{bin_fpga} designed to run neural networks on edge devices at the cost of some performance loss.
\par Currently, hardware development is moving past only targeting edge devices and into the server farm, where deep learning architectures such as transformers have become so massive, slow, and energy intensive as to demand custom hardware solutions.  This has led to cutting edge accelerators such as the Google TPU \cite{tpu}, Cerebras CS-2 \cite{cerebras}, and Groq TSP \cite{groq} among others.  This development opens the door to new approaches in algorithm development. Conventionally, binarization techniques would have been focused on lightweight edge devices, making long context applications unrealistic.  Additionally, long-context work would be constrained to target conventional GPU hardware, which is optimized for dense, floating point arithmetic.  Our work bridges this gap, selectively applying binarization techniques to target custom hardware for efficient long context inference.  Our main contributions are as follows:
\begin{enumerate}
    \item We present a novel, fast, and lightweight framework for fine-tuning transformers to enable binarization of keys and queries and sparsification of the attention matrix, accelerating all $O(n^2)$ attention operations.
    \item We evaluate the proposed binarized models across various architectures and tasks, showcasing minimal accuracy degradation.
    \item We assess the performance of a binarized model on the QuALITY long-context question answering benchmark \cite{quality}, demonstrating improved performance with extended context.
    \item We simulate the computational efficiency of our binarized attention mechanism on custom hardware, revealing significant performance gains compared to standard attention.
\end{enumerate}

\section{Related Work}

\subsection{Long Context Architectures}
A wide range of variations of and replacements for self-attention have been proposed to address the $O(n^2)$ scaling problem.  A number of these approaches use fixed, sparse attention maps, with fewer than $O(n^2)$ active elements.  For example, longformer \cite{longformer} combines local dilated and non-dilated sliding-window attention with a fixed number of globally attending tokens, producing an attention matrix with $O(n)$ non-zero elements and fully networked tokens over multiple sequential attention layers.  BigBird \cite{bigbird} is similar, but also includes randomly selected non-zero elements in the attention matrix. \par Another approach, the Reformer \cite{reformer}, uses locality sensitive hashing to bucket keys and queries, and performs attention within these limited buckets.  All of these approaches produce non-trivial accuracy losses against otherwise equivalent models.  More recently, Mamba \cite{mamba} has gained popularity as a competitive attention replacement across many tasks, radically departing from attention and using a state space model instead.  While promising, this offers a radically different set of tradeoffs from standard attention, and is even more poorly suited as a drop-in replacement to finetune pre-trained transformers for long context tasks.  
\par Across the long-context field, almost all works focus on asymptotic runtime improvements rather than improving hardware utilization, and very few consider adapting pre-trained transformers.

\subsection{Deep Neural Network Binarization}
Binarized neural networks \cite{bnn}, trained with binary weights and activations, were developed to increase the efficiency of deep neural network inference, and predate the transformer architecture.  While early works focused on convolutional neural networks \cite{bin_cnn}, many recent works have adapted binarization techniques to address the transformer architecture.  Some of these works use variations of weight binarization while quantizing activations gently or not at all \cite{pbllm,onebit}.  This achieves very small performance degredation at the cost of increased runtime and memory when compared to full binarization approaches. 
\par These approaches are well suited for decoder-only GPT-style language models, where the quality of generated text is highly sensitive to small changes in the loss value, scaling the number of parameters drastically improves performance, and long contexts are not often necessary.  However, these approaches do little to nothing to address the computational costs of attention which dominates in long-context settings, as attention is an operation only between activations in the form of keys, queries, and values.  
\par Other works, generally applied to BERT-style encoder-only models and vision transformers, perform full binarization including activations and attention, maximizing efficiency at the cost of accuracy.  Among the state of the art in this category is BiViT \cite{bivit}, which proposes a novel softmax-aware attention matrix binarization function, achieving significantly higher accuracies than previous fully binarized vision transformers.  BiT \cite{bit} achieved state-of-the-art results on the GLUE benchmark among fully binarized BERT variants, learning scale and threshold terms during binarization and using a multi-step distillation procedure.  Much like BiT, our work includes explicit loss terms so that the fine-tuned student model's attention map closely replicates the teacher model's.  Effectively all of the above listed works use variations of straight-through-estimators (STEs) to estimate gradients for non-differentiable quantized functions, but HAD also uses transformations of the hyperbolic tangent (tanh) function to smoothly transition from continuous to binarized representations, similar to previous binarization works \cite{tanh_bin}. 
\par Many highly optimized binary multipliers exist, which are tailored to perform XNOR-based computations that reduce area and power while maintaining performance \cite{rastegari2016xnor,lee2022power,garg2013array}. These multipliers leverage the inherent simplicity of binary operations to enable faster computation with reduced hardware requirements, therefore accelerating binarized neural networks. We would expect binarized attention activations to be particularly valuable, as processing in memory approaches which have been successful in accelerating weight-activation operations \cite{pim} are not as easily applied to activation-activation operations in attention, exacerbating its costs.

\subsection{Binary Keys and Queries}
The selective binarization of keys and queries in this work is primarily inspired by associative memory and Hopfield networks.  In hopfield networks, a set of binary vectors or "patterns" are stored and can be queried according to some update rule.  Modern hopfield networks \cite{modern_hop} are capable of addressing a number of unique patterns scaling exponentially with the number of bits per pattern. "Hopfield networks is all you need" \cite{hopfield_is} demonstrated that self-attention is effectively a generalization of modern hopfield networks to continuous valued patterns in the forms of keys, addressed by queries.  This mapping between attention and binary modern hopfield networks, along with the exponential storage capacity of modern hopfield networks, led us to hypothesize that binarizing specifically our keys and queries would still enable expressive attention lookups to retrieve value vectors.  \par Additionally, studies of scaling laws of transformer-based LLMs find that performance is tightly coupled to parameter count and training data, and very weakly to model shape \cite{scaling}, indicating that the capacity of the attention mechanism does not determine the capacity of the transformer in many cases.  Therefore, we do not expect the capacity loss induced by key and query binarization to signficantly reduce model performance.

\subsection{Sparse Attention}
Like multiple other works, HAD utilizes a sparse attention matrix.  Because the attention matrix is the output of a softmax function, and the relative magnitude of outputs of a softmax is exponential relative to the difference in inputs, large attention matrices contain a large number of near-zero values.  Some works such as BigBird \cite{bigbird} and Longformer \cite{longformer} impose structured sparsity on the attention matrix, with a bias towards nearby elements being able to attend to one another.  Other works, exploit the intrinsic sparsity of the attention matrix in order to filter out unwanted features \cite{sparse_att_filter} or accelerate transformer inference \cite{sparse_att_acc} like HAD.  
\par We select the top $N$ highest attention scores corresponding to each query, and performing a sparse accumulation over our values matrix.  While binarization allows us to accelerate the $O(n^2)$ $K Q^T$ operation, sparsity allows us to accelerate the $O(n^2)$ softmax, scaling, and accumulation over $V$. Many works have proposed GPU and custom hardware solutions, leveraging sparsity to improve efficiency \cite{sparse_asic,sparse_gpu}.

\section{Methods}

\subsection{Overview}

\begin{figure}[t]
    \centering
    \includegraphics[width=0.6\linewidth]{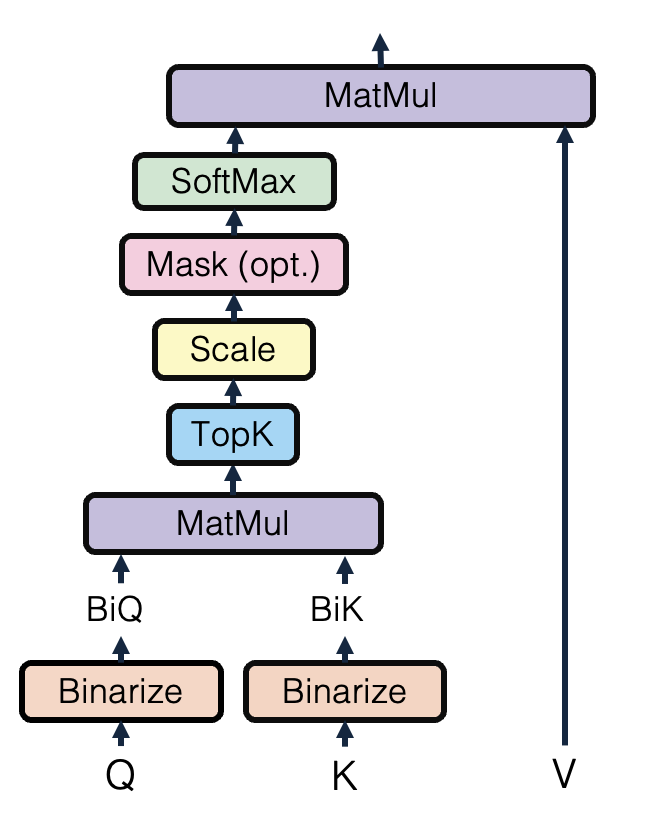}
    \caption{Binarized attention mechanism in Hamming Attention Distillation (HAD), illustrating the binarization of keys (K) and queries (Q) and subsequent attention operations.}
    \label{fig:attention}
\end{figure}

To improve the efficiency of self-attention in long-context scenarios, we propose a novel distillation approach such that given a pre-trained transformer, we can binarize the query ($Q$) and key ($K$) matrices and sparsify the attention matrix ($A$).  Our method retains the top $N$ elements of $A$ per query vector, retaining the most relevant attention interactions.  We can describe standard attention with the following equations where $K$, $Q$, and $V$ are our key, query, and value matrices respectively, $A_l$ is our attention logit matrix,  $A$ is our attention matrix, and $d_k$ is the dimension of our attention head:
\begin{equation}
    A_l = \frac{QK^\top}{\sqrt{d_k}}
\end{equation}
\begin{equation}
    A = \text{softmax}(A_l)
\end{equation}
\begin{equation}
    \text{Output} = AV
\end{equation}

In contrast, our modified attention mechanism diagrammed in figure \ref{fig:attention} can be described by the following equations where $Q_c$ and $K_c$ are our continuous-valued query and key matrices prior to binarization:

\begin{equation}
    \text{Q} = \text{sign}(Q_c), 
    \quad
    \text{K} = \text{sign}(K_c)
\end{equation}
\begin{equation}
    A_l = Q \cdot K^\top
\end{equation}
\begin{equation}
    A_{\text{topn}} = \text{topn}(A_l, N)
\end{equation}
\begin{equation}
    A = \text{softmax}(A_{\text{topn}} / \sqrt{d_k})
\end{equation}
\begin{equation}
    \text{Output} = AV
\end{equation}

In order to avoid computation on unused attention elements, we perform scaling and softmax and  after our top $N$ operation.  In order to achieve this binarization, we implement the following multi-stage distillation algorithm.  

\begin{algorithm}[H]
\caption{Proposed Training Algorithm}
\label{alg:training}
\begin{algorithmic}[1]
    \REQUIRE Pre-trained teacher model $M_T$
    \ENSURE Optimized student model $M_S$
    \STATE \textbf{Initialize:} Copy teacher model weights to student model $M_S \gets M_T$
    \STATE Calculate standardization coefficients $\sigma_Q, \sigma_K$
    \STATE \textbf{Stage 1:} Train with scaled tanh approximation, decreasing $c$ from 5 to 1, with $A_l$ distillation loss
    \STATE \textbf{Stage 2:} Train with tighter tanh approximation, decreasing $c$ from 1 to 0.05, with $A_l$ distillation loss
    \STATE \textbf{Stage 3:} Apply STE-based binarization with $A_l$ distillation loss
    \STATE \textbf{Stage 4:} Fine-tune at a lower learning rate without $A_l$ distillation loss
    \RETURN Optimized student model $M_S$
\end{algorithmic}
\end{algorithm}

In the following sections, we provide a detailed explanation of our approach. Specifically, we cover:
\begin{itemize}
    \item \textbf{Selecting N:} We detail our procedure for selecting our sparsity parameter $N$.
    \item \textbf{Loss functions:} We define the loss components used during training, including distillation loss and standard training objectives.
    \item \textbf{Standardization coefficients:} A description of the computation of $\sigma_Q$ and $\sigma_K$ and their role in binarization.
    \item \textbf{Training stages:} A breakdown of each of the four training stages, detailing their objectives and transition criteria.
\end{itemize}

\subsection{Selecting N}
\begin{figure}[t]
    \centering
    \includegraphics[width=0.9\linewidth]{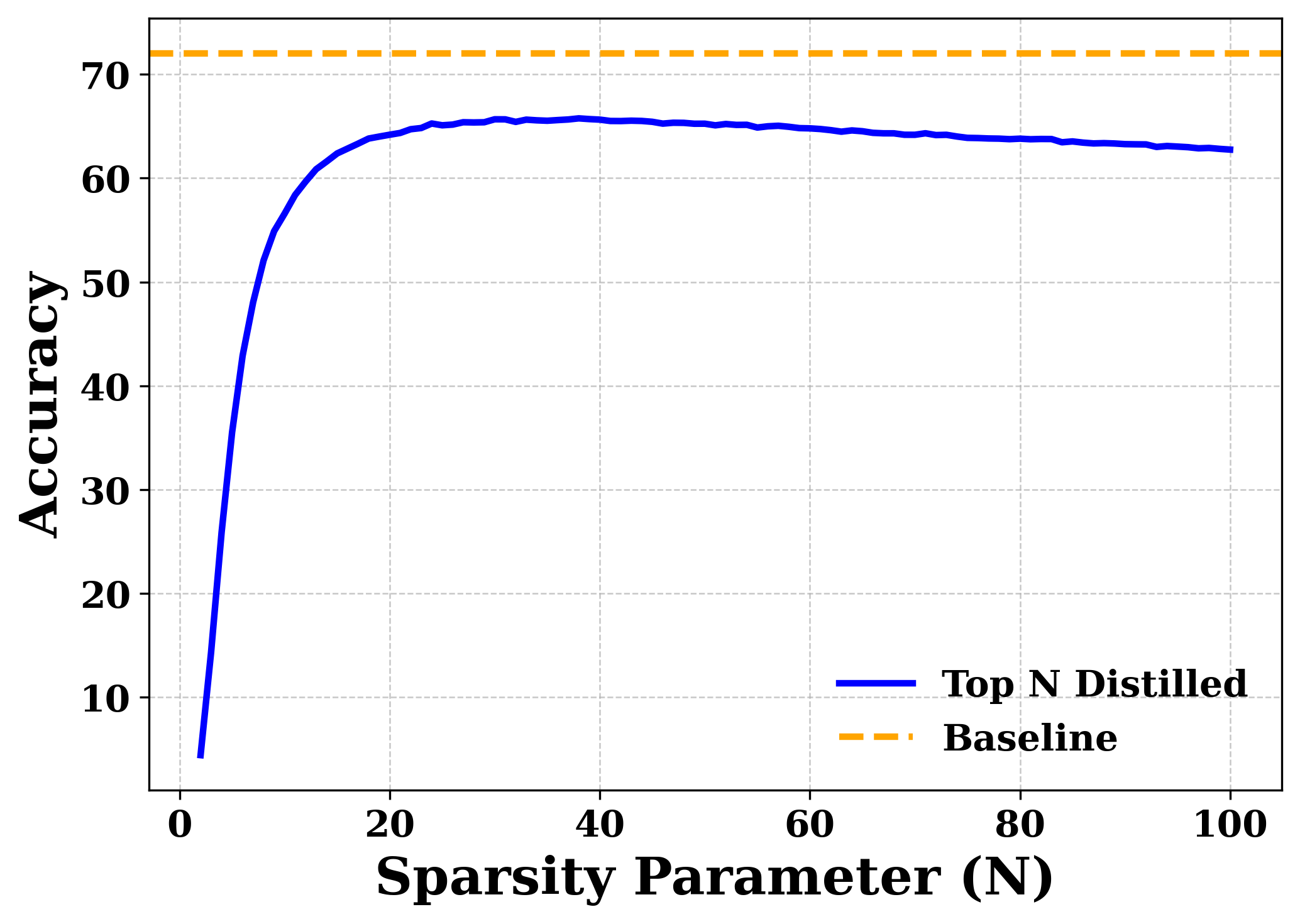}
    \caption{Accuracies measured while progressively distilling a full precision DeiT-T over decreasing N values.}
    \label{fig:n_selection}
\end{figure}

\begin{figure}[t]
    \centering
    \includegraphics[width=1.0\linewidth]{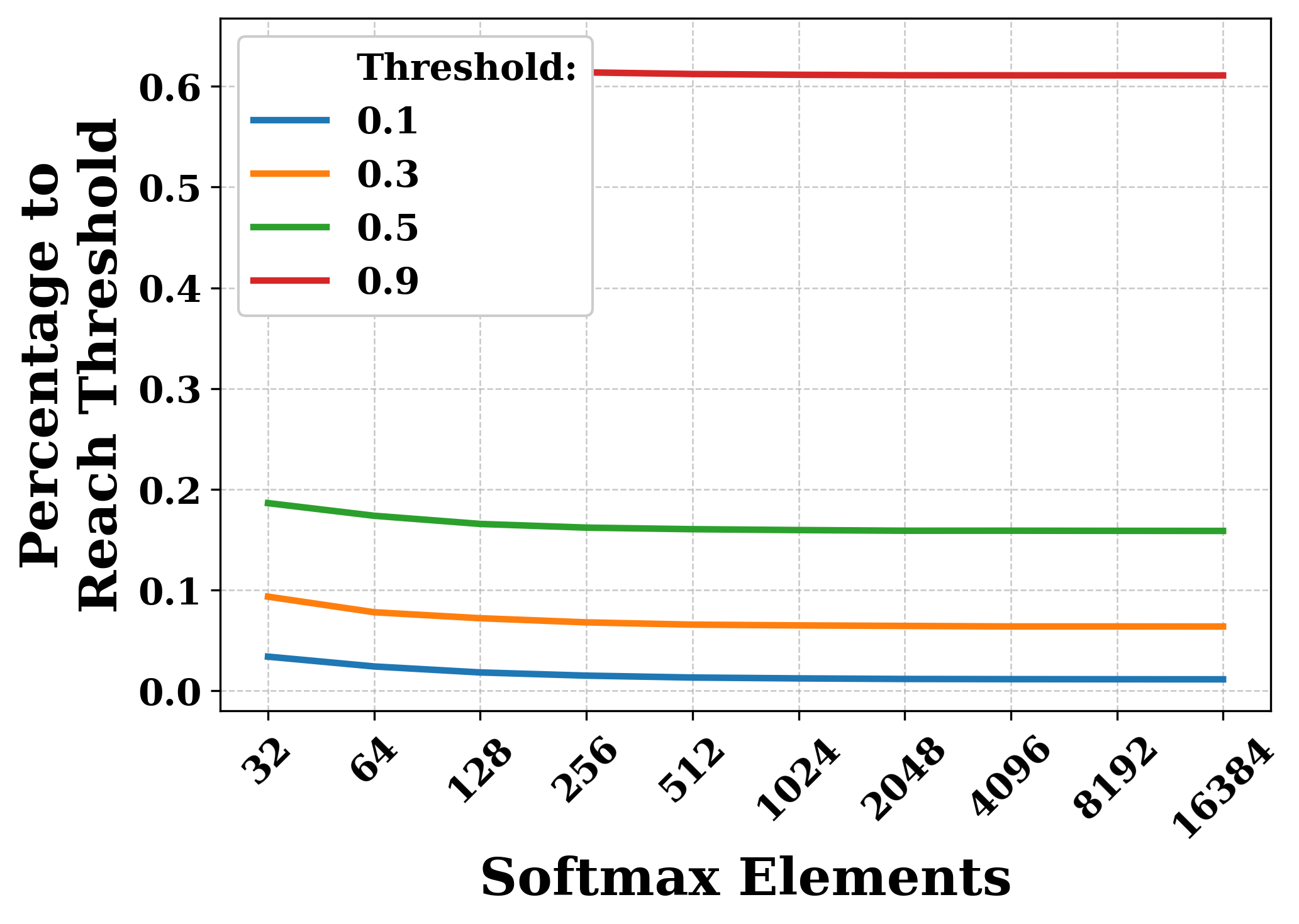}
    \caption{Given standard gaussian inputs, the percentage of the largest softmax outputs required to sum to the threshold probability.  In effect, how many elements are required to account for some percentage of probability mass.}
    \label{fig:softmax_scaling}
\end{figure}

Before we start, we must decide how to set our sparsity parameter $N$.  Figure \ref{fig:n_selection} shows our accuracies distilling a full-precision DeiT tiny model \cite{deit} with top $N$ sparsity from its teacher.  Starting at $N=100$ and gradually decreasing, we see accuracy recovery until about $N=30$ at which point it starts decreasing.  DeiT has a context length of 197, and we fine that $N=30$ also works well for our BERT models (Section \ref{subsec:glue}) \cite{bert} with a similar context length of 256 as well as our DeiT models (Section \ref{subsec:imagenet}).  However, as we scale to longer contexts, we look to Figure \ref{fig:softmax_scaling} for insights.  We find that over larger softmax operations, the percentage of the largest outputs needed to add to some constant probability threshold approaches a constant.  We use this as a guiding principle, linearly scaling $N$ with context length in Section \ref{subsec:quality} 

\subsection{Loss Functions}
To maintain accuracy after binarization and sparsification, we employ a teacher-student distillation approach. Let \( A_{l,t}^{(m)} \) and \( A_{l,s}^{(m)} \) represent the attention logit matrices from the teacher and student models, respectively, for attention matrix \( m \). We define the Kullback–Leibler (KL) divergence loss for the attention logits as the unweighted mean over all rows of all attention heads:

\begin{equation}
\label{eq:kl_att}
\begin{gathered}
    \mathcal{L}_{\text{KL-att}} = \\
    \frac{1}{Mn}
    \sum_{m=1}^{M}
    \sum_{i=1}^{n}
    \sum_{j=1}^{n}
    \exp(A_{l,t}^{(m)}(i,j))
    \left(
        A_{l,t}^{(m)}(i,j) - A_{l,s}^{(m)}(i,j)
    \right)
\end{gathered}
\end{equation}

where \( M \) represents the total number of attention maps (\(\text{heads} \cdot \text{layers}\)), and \( n \) represents the number of rows (queries) in each attention map. This formulation encourages the binarized queries and keys, \( Q \) and \( K \), to produce attention maps approximating those of their full precision counterparts.

Additionally, we incorporate a KL divergence loss on the model’s output logits, which is defined as:

\begin{equation}
\label{eq:kl_out}
    \mathcal{L}_{\text{KL-out}} 
    =
    \sum_{i}
    \exp(z_{t,i}) 
    \left(
        z_{t,i} - z_{s,i}
    \right)
\end{equation}

where \( z_t \) and \( z_s \) are the output logits of the teacher and student models, respectively.

Our final training objective is the sum of the attention and output distillation losses:
\begin{equation}
    \mathcal{L} 
    \;=\;
    \mathcal{L}_{\text{KL-att}}
    \;+\;
    \mathcal{L}_{\text{KL-out}}.
\end{equation}

\subsection{Standardization Coefficients}
To facilitate the binarization of the query and key matrices, we estimate the standard deviations \( \sigma_Q \) and \( \sigma_K \) based on the training data.

Our estimation procedure involves performing inference on a subset of the training data. Specifically, we randomly sample 100 minibatches, each containing 16 samples, from the training dataset. For each layer in the model, the standard deviation is computed independently within each minibatch and averaged over all minibatches.

Let \( Q^{(b)} \) and \( K^{(b)} \) represent the query and key matrices for minibatch \( b \), respectively. The standard deviation for each minibatch is computed as follows:
\begin{equation}
    \sigma_Q = \frac{1}{100} \sum_{b=1}^{100} \text{std}(Q^{(b)}), \quad
    \sigma_K = \frac{1}{100} \sum_{b=1}^{100} \text{std}(K^{(b)}),
\end{equation}

where \( \text{std}(\cdot) \) represents the standard deviation operation applied across all elements within the corresponding matrix.  
\par We find empirically that models such as T5 \cite{t5} with $K$ and $Q$ standard deviations far larger or smaller than 1 require standardization with these coefficients for good binarized performance.

\subsection{Stage 1: Approaching Tanh}

In the first stage of training, we aim to approach a tanh activation for $Q$ and $K$ matrices while preserving essential attention patterns. This is achieved by gradually reducing the scaling parameter \( c \) in the binarization approximation function. During stage 1, we apply top \( N \) masking to sparsify the attention matrix and employ our combined loss function to guide the training process.

We approximate the binarization of the query and key matrices by using a scaled tanh function. Specifically, the query and key matrices are transformed as follows:

\begin{equation}
    Q = c \cdot \sigma_Q \cdot \tanh\left(\frac{Q_c}{c \cdot \sigma_Q}\right), \quad
    K = c \cdot \sigma_K \cdot \tanh\left(\frac{K_c}{c \cdot \sigma_K}\right),
\end{equation}

At high c values, this function behaves similar to a linear activation, and at $c=1$ it behaves similar to tanh scaled by $\sigma$.  We decay $c$ exponentially from 5.0 until it reaches 1.0, at which point we begin stage 2.

\subsection{Stage 2: Approaching Sign}

In the second stage of training, we aim to approach a sign activation for $Q$ and $K$ by further reducing the scaling parameter \( c \) in the binarization approximation function. Like in stage 1, we apply top $N$ masking and out combined loss function.  Our stage 2 transformations are as follows:

\begin{equation}
    Q = c \cdot \sigma_Q \cdot \tanh\left(\frac{Q_c}{c \cdot \sigma_Q}\right), \quad
    K = c \cdot \sigma_K \cdot \tanh\left(\frac{K_c}{c \cdot \sigma_K}\right),
\end{equation}

We approximate the binarization of the query and key matrices by using a scaled tanh function. Specifically, the query and key matrices are transformed as follows:

\begin{equation}
    Q = \sigma_Q \cdot \tanh\left(\frac{Q_c}{c \cdot \sigma_Q}\right), \quad
    K = \sigma_K \cdot \tanh\left(\frac{K_c}{c \cdot \sigma_K}\right),
\end{equation}

At c=1 this function is equivalent to the end of stage 1, but as it approaches 0 this function approaches the sign function scaled by $\sigma$.  We continue to decay $c$ exponentially from 1.0 until it reaches 0.05, at which point we begin stage 3.

\subsection{Stage 3: Straight-Through-Estimator Training}
In stage 3, we aim to refine our binarized $K$ and $Q$ by training with a STE.  Once again, we continue to use top $N$ masking and the combined loss.  We use the following STE function:

\begin{equation}
    \text{STE}_{\text{forward}}(x) = \text{sign}(x),
\end{equation}
\begin{equation}
    \frac{\partial \text{STE}_{\text{forward}}(x)}{\partial x} =
    \begin{cases}
        1, & -1 \leq x \leq 1 \\
        0, & \text{otherwise}
    \end{cases}
\end{equation}

which results in the following $K$ and $Q$ transformations:

\begin{equation}
    Q = \sigma_Q \cdot \text{STE}\left(\frac{Q_c}{\sigma_Q}\right), 
    \quad
    K = \sigma_K \cdot \text{STE}\left(\frac{K_c}{\sigma_K}\right)
\end{equation}

We run this for 10,000 iterations before moving on to stage 4.

\subsection{Stage 4: Final Refinement}
In stage 4, we use the same binarization function and STE as stage 3.  However, we now lower the learning rate and remove the attention map distillation loss, allowing the model more flexibility to adjust its attention maps in service of the output distillation loss:

\begin{equation}
    \mathcal{L}(z_t, z_s) = \sum_{i} \exp(z_{t,i}) \left( z_{t,i} - z_{s,i} \right)
\end{equation}

After 10,000 iterations, we terminate.

\subsection{Training Details}
All models are distilled with a batch size of 16, the Adam optimizer, a learning rate of $10^{-5}$ during stages 1-3 and $10^{-6}$ during stage 4, gradient clipping at magnitude 0.5, and a c decay coefficient of $0.9998$ per minibatch.

\section{Results and Discussion}

\subsection{GLUE}
\label{subsec:glue}
\begin{table*}[t]
    \centering
    
    \caption{GLUE benchmark results comparing our method to the Hugging Face BERT baseline models from TextAttack, BiT, and several ablation studies. "w/ SAB" includes softmax-aware attention map binarization as described in BiViT \protect\cite{bivit}. "w/o AD" excludes attention map distillation, and "w/o tanh" removes the tanh training stage, replacing it with an equivalent number of STE training steps.}
    \begin{tabular}{lr:rr:rrr}
                \toprule
        Benchmark & Baseline & HAD (ours) & BiT & w/ SAB & w/o AD & w/o Tanh \\
        \midrule
        MNLI   & 84.58/84.46 & \textbf{82.45/82.84} & 79.5/79.4  & 55.31/56.04  & 82.00/82.25  & 82.23/82.58  \\
        QQP    & 90.91  & 90.11  & 85.4  & 78.34  & 89.51  & \textbf{90.15}  \\
        QNLI   & 91.54  & \textbf{89.68}  & 86.4  & 71.24  & 89.75  & 89.64  \\
        SST-2  & 92.43  & \textbf{91.63}  & 89.9  & 82.34  & 90.71  & 90.60  \\
        CoLA   & 53.39  & 55.47  & 32.9  & 16.41  & \textbf{55.82}  & 52.32  \\
        STS-B  & 87.63  & \textbf{87.46}  & 72.0  & 38.70  & 86.77  & 86.78  \\
        MRPC   & 87.75  & \textbf{83.82}  & 79.9  & 68.87  & 82.11  & 83.5  \\
        RTE    & 72.56  & 65.70  &  62.1  & 49.82  & 64.26  & \textbf{66.06}  \\
        \hdashline
        \textbf{Avg}    & 82.59  & \textbf{80.81}  &  73.51  & 57.67  & 80.13  & 80.19  \\
        \bottomrule
    \end{tabular}
    \label{tab:glue_res}
\end{table*}

Table \ref{tab:glue_res} presents results for BERT backbone models distilled and evaluated using the GLUE language understanding benchmark, comparing our approach to BiT \cite{bit}, which employs full binarization, along with several ablation studies. We observe that by binarizing only the Q and K activations while keeping all other weights and activations in full precision, our method consistently achieves significantly higher accuracies compared to BiT's full binarization approach. All models used a maximum context length of 256 tokens and all HAD variants used top 30 sparsification of the attention vector per query.  All methods seem to significantly struggle with RTE and MRPC, but on all other tasks HAD achieves accuracies within 3\% of the baseline teacher model.  Notably, we did not face noticeable overfitting issues on any of these tasks, despite running hundreds of distillation epochs for some of the smaller benchmark training datasets.
\par To further assess the impact of binarizing K and Q versus the attention matrix A, we incorporated the softmax-aware attention binarization (SAB) function and the Straight-Through Estimator (STE) from BiViT \cite{bivit}, while keeping the rest of the training pipeline unchanged ("w/ SAB"). This resulted in substantial accuracy losses. However, this does not necessarily imply that architectures utilizing SAB cannot achieve competitive performance. Rather, it suggests that systems with binarized A may struggle to close the accuracy gap with those using full-precision A and may be highly sensitive to factors such as batch size, training time, and learning rate.
\par Additionally, we conducted ablation studies by excluding our attention map distillation loss ("w/o AD") and the tanh training phase ("w/o Tanh"), replacing these with an equivalent number of STE training iterations. We find that both ablations perform comparably to our standard training procedure across many tasks, but yield non-trivial performance losses in some cases.  It is worth noting, we observed that attention distillation and tanh binarization contribute significantly when using less optimized training pipelines. We find these techniques add robustness against variations in architectures, tasks, the absence of $\sigma_K$ and $\sigma_Q$ normalization, and hyperparameters such as learning rate, batch size, training time, and gradient clipping.

\subsection{ImageNet}
\label{subsec:imagenet}
\begin{table}[t]
    \centering
    \caption{ImageNet benchmark results comparing our method to the Hugging Face DeiT baseline models from Facebook Research, BiViT, and several ablation studies. "w/ SAB" includes softmax-aware attention map binarization as described in BiViT \protect\cite{bivit}. "w/o AD" excludes attention map distillation, and "w/o Tanh" removes the tanh training stage, replacing it with an equivalent number of STE training steps.}

    \begin{tabular}{lrr}
        \toprule
        & DeiT-B & DeiT-T \\
        \midrule
        Baseline  & 81.74  & 72.01  \\
        HAD (ours)  & 79.24  & 66.59  \\
        BiViT  & 69.6  & 37.9  \\
        w/ SAB  & 6.36  & 4.32  \\
        w/o AD  & 79.29  & 66.42  \\
        w/o Tanh  & \textbf{79.52}  & \textbf{66.78}  \\
        \bottomrule
    \end{tabular}
    \label{tab:imagenet_res}
\end{table}

Table \ref{tab:imagenet_res} presents results for the base and tiny variants of DeiT \cite{deit}, distilled and evaluated on the ImageNet \cite{imagenet} image classification benchmark.  We once again find that the selective binarization of Q and K produces significantly higher accuracies than implementations including binarization of the attention matrix, as we also saw with the GLUE evaluations.  We find that all binarization techniques struggle with the tiny variant of DeiT, indicating that very low capacity attention modules may be disproportionately adversely effected by binarization and/or top $N$ sparsification.  When evaluating on ImageNet, the ablations were on par with our full fine-tuning procedure, indicating that these models are less sensitive to attention map distillation and gradual binarization using the STE.

\subsection{Long Context Evaluation}
\label{subsec:quality}
\begin{figure}[t]
    \centering
    \includegraphics[width=1.0\linewidth]{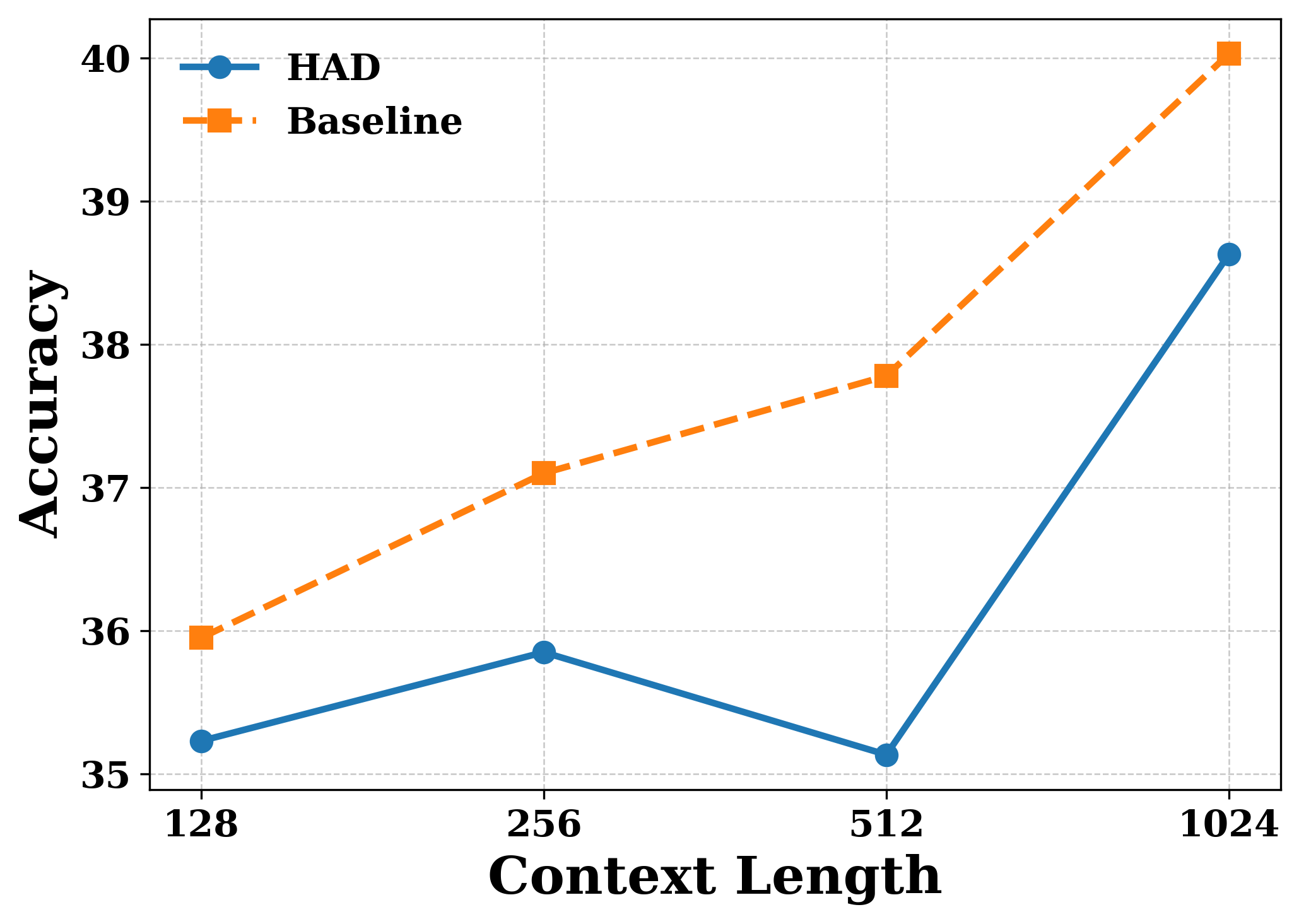}
    \caption{Comparison of HAD and baseline accuracy across different context lengths, evaluated on QuALITY \protect\cite{quality}. The context lengths are powers of 2, and the accuracy is measured for both methods.}
    \label{fig:quality}
\end{figure}
To evaluate HAD's performance in long-context settings, we trained and tested models using the QuALITY long-context multiple-choice question-answering dataset \cite{quality} at various context lengths. The QuALITY benchmark contains inputs ranging from 2,000 to 6,000 words, which we truncated to fit the context limits of our models. For our baseline, we used Google's T5-Base model \cite{t5}, which was pre-trained on the RACE dataset \cite{race} and fine-tuned on QuALITY truncated to each context length. These truncated QuALITY trainind data were also used for our distillation dataset.
\par While our token limits were set to 256 for GLUE and 197 for ImageNet, here we explore context lengths from 128 to 1024 tokens in powers of two. The top $N$ sparsification parameter $N$ was scaled linearly with context length, ranging from 15 at 128 tokens to 120 at 1024 tokens, ensuring a consistent sparsity percentage across experiments.
\par We observed some noise in accuracies across runs and epochs, which accounts for the performance drop at 512 tokens. However, overall, HAD’s accuracy followed the baseline model’s trend of improvement with increasing context length, achieving within 3\% of the baseline model’s accuracy. Combined with our earlier analysis of the softmax function in the long-context regime, these findings support the conclusion that HAD effectively scales to long contexts while maintaining strong performance.

\subsection{Hardware Optimization through Algorithmic Design}

To evaluate the impact of algorithmic optimizations on hardware, we synthesized and analyzed a CAM-based binary attention architecture and a conventional BF16 attention design. 
HAD's custom design leverages capacitive Content-Addressable Memory (CAM) to replace traditional matrix multiplication in attention mechanisms with in-memory associative matching. Its architecture integrates 1-bit XNOR operations for binarized query-key vectors and employs top \(N\) sparsity to significantly reduce power and area.
In contrast, the BF16 digital attention showed significantly higher area and power metrics: 31.795~mm\textsuperscript{2} and 25.491~W.
Both designs performed attention computation via QK multiplication for (1×1024) × (1024×256) and Attention Probabilities × V for (1x256) × (256x1024).
Power and area were estimated using Verilog for a smaller module, synthesized with Synopsys Design Compiler, and scaled to the full design.

\begin{table}[t]
\scriptsize
\centering
\caption{Comparison of Area and Power between Standard Attention (SA) and HAD hardware implementations of an attention head with a context length of 256 and $N=30$. HAD achieves ~79\% area and ~87\% power reduction via CAM-based XNOR operations and sparsity, while maintaining competitive accuracy, highlighting its efficiency over standard attention. }
\begin{adjustbox}{width=0.9\columnwidth,center}
\begin{tabular}{lrr|rr}
    \toprule
    Component & \multicolumn{2}{c|}{\textbf{Area (mm\(^2\))}} & \multicolumn{2}{c}{\textbf{Power (W)}}  \\
    \cmidrule(r){2-3} \cmidrule(r){4-5}
              & \textbf{SA} & \textbf{HAD} & \textbf{SA} & \textbf{HAD} \\
    \midrule
    Q K       & 15.880 & 1.108 & 12.730 & 0.127     \\
    Top $N$     & 0.000  & 0.008 & 0.000  & 0.009  \\
    SoftMax   & 0.035  & 0.017 & 0.031  & 0.024 \\
    A V       & 15.880 & 5.591 & 12.730 & 3.141  \\
    Total     & 31.795 & 6.724 & 25.491 & 3.301 \\
    \bottomrule
\end{tabular}
\end{adjustbox}
\label{tab:area_power_comparison}
\normalsize
\end{table}

\section{Conclusion}
In this paper, we presented Hamming Attention Distillation (HAD), a novel distillation framework applying selective binarization to address long context transformer inference.  HAD selectively binarizes the key (\(K\)) and query (\(Q\)) projections in transformers while sparsifying the attention matrix. By reducing floating point operations, this approach enables highly efficient hardware deployment with a very small performance cost when handling long context inputs. Through distillation, HAD retains strong performance on benchmarks such as GLUE, ImageNet, and QuALITY, while using a fraction of the compute resources of standard attention.  

Future work may explore extending our method to efficient GPU implementations, increasing tolerated sparsity, and reducing the precision of the $V$ matrix to further accelerate the $A V$ value accumulation operation.  Additionally, these methods may be adapted to decoder-only LLMs, which are very sensitive to small loss increases and do not tolerate general activation binarization.  Finally, variations of this work could be explored as train-time optimizations, using binarization and sparsity to reduce the overhead of training transformer models on long sequences.

\FloatBarrier

\bibliographystyle{plain}
\bibliography{ijcai25}

\end{document}